\documentclass{article}

\usepackage{threeparttable}
\usepackage{times}
\usepackage{epsfig}
\usepackage{graphicx}
\usepackage{amsmath}
\usepackage{amssymb}
\usepackage[english]{babel}
\usepackage{pdfrender}

\usepackage{hyperref}
\usepackage{url}
\usepackage{cleveref}
\hypersetup{colorlinks}
\usepackage{mathtools}
\usepackage{algorithm}
\usepackage{dpr}
\usepackage{fec}
\usepackage{enumerate}
\usepackage{subcaption}
\usepackage{makecell}
\usepackage{comment}
\usepackage{caption}
\usepackage{subcaption}
\usepackage{enumitem}
\usepackage{amsthm}
\usepackage{arydshln}
\usepackage{bm}
\usepackage{siunitx,tabularx,ragged2e,booktabs}
\usepackage{wrapfig}
\usepackage{float}
\usepackage{booktabs}
\usepackage{pifont}
\usepackage{soul}
\usepackage{epstopdf}
\usepackage{blindtext}
\usepackage{fancyvrb}
\usepackage{multirow, booktabs}
\usepackage{cleveref}
\usepackage{hhline}

\newcommand{\algacro}{LoRAShear{}}
\newcommand{\lora}{LoRA{}}
\newcommand{\llm}{LLM{}}

\newcommand{\lorahspg}{\text{LHSPG}}

\newcommand{\llamavone}{\text{LLAMAv1}}

\newcommand{\ie}{\textit{i.e.}}
\newcommand{\eg}{\textit{e.g.}}

\newcommand{\xcheckmark}{\checkmark\kern-1.1ex\raisebox{.7ex}{\rotatebox[origin=c]{125}{--}}}
\usepackage[accepted]{icml2023}

\usepackage{amsmath}
\usepackage{amssymb}
\usepackage{mathtools}
\usepackage{amsthm}
\usepackage{siunitx}
\usepackage{xcolor}

\theoremstyle{plain}
\theoremstyle{definition}
\theoremstyle{remark}

\usepackage[textsize=tiny]{todonotes}

\usepackage[outline]{contour}
\contourlength{0.25pt}
\contournumber{20}

\usepackage{framed}
\usepackage{tikz}
\usetikzlibrary{arrows}
\usepackage{setspace}
\tikzset{
  treenode/.style = {align=center, inner sep=0pt, text centered,
    font=\sffamily},
  arn_n/.style = {treenode, circle, white, font=\sffamily\bfseries, draw=black,
    fill=black, text width=1.5em},% arbre rouge noir, noeud noir
  arn_r/.style = {treenode, circle, red, draw=red,
    text width=1.5em, very thick},% arbre rouge noir, noeud rouge
  arn_x/.style = {treenode, rectangle, draw=black,
    minimum width=0.5em, minimum height=0.5em}% arbre rouge noir, nil
}

\fvset{frame=single,framesep=1mm,fontfamily=courier,fontsize=\scriptsize,numbers=left,framerule=.3mm,numbersep=1mm,commandchars=\\\{\}}

\title{xxx}

\icmltitlerunning{Preprint version.}

\begin{document}

\twocolumn[
\icmltitle{\algacro{}: Efficient Large Language Model Structured Pruning and Knowledge Recovery}

% \icmlsetsymbol{equal}{*}

\begin{icmlauthorlist}
	\icmlauthor{Tianyi Chen}{comp}
	\icmlauthor{Tianyu Ding}{comp}
	\icmlauthor{Badal Yadav}{comp}
	\icmlauthor{Ilya Zharkov}{comp}
	\icmlauthor{Luming Liang}{comp}\\
	$^1$Microsoft\\
	\{tiachen,tianyuding,bayadav,zharkov,lulian\}@microsoft.com
\end{icmlauthorlist}

\icmlaffiliation{comp}{Microsoft, Redmond WA 98052, United States}

\icmlcorrespondingauthor{Tianyi Chen}{tiachen@microsoft.com}
\icmlcorrespondingauthor{Luming Liang}{lulian@microsoft.com}

% You may provide any keywords that you
% find helpful for describing your paper; these are used to populate
% the "keywords" metadata in the PDF but will not be shown in the document
\icmlkeywords{Machine Learning, ICML}

\vskip 0.3in
]

% this must go after the closing bracket ] following \twocolumn[ ...

% This command actually creates the footnote in the first column
% listing the affiliations and the copyright notice.
% The command takes one argument, which is text to display at the start of the footnote.
% The \icmlEqualContribution command is standard text for equal contribution.
% Remove it (just {}) if you do not need this facility.

\printAffiliationsAndNotice{}  % leave blank if no need to mention equal contribution
% \printAffiliationsAndNotice{\icmlEqualContribution} % otherwise use the standard text.

\begin{abstract}
Large Language Models (LLMs) have transformed the landscape of artificial intelligence, while their enormous size presents significant challenges in terms of computational costs. We introduces \algacro{}, a novel efficient approach to structurally prune LLMs and recover knowledge. Given general LLMs, \algacro{} at first creates the dependency graphs over LoRA modules to discover minimally removal structures and analyze the knowledge distribution. It then proceeds progressive structured pruning on \lora{} adaptors and enables inherent knowledge transfer to better preserve the information in the redundant structures. To recover the lost knowledge during pruning, \algacro{} meticulously studies and proposes a dynamic fine-tuning schemes with dynamic data adaptors to effectively narrow down the performance gap to the full models.  Numerical results demonstrate that by only using one GPU within a couple of GPU days, \algacro{} effectively reduced footprint of LLMs by 20\% with only 1.0\% performance degradation and significantly outperforms state-of-the-arts. (Code will be public soon.)
\end{abstract}

\section{Introduction}

The advent of Large Language Models (LLMs)~\citep{zhao2023survey, hadi2023survey} has marked a significant milestone in evolution of artificial intelligence. These models, distinguished by their extensive parameter sizes, have demonstrated emergent abilities~\citep{wei2022emergent}, catalyzing breakthroughs not only in the realm of natural language processing but also across tasks in various domains~\citep{driess2023palm}. This has opened up new possibilities for advancing towards Artificial General Intelligence~(AGI)~\citep{everitt2018agi,bubeck2023sparks}. However, the enormous size of LLMs, typically ranging from tens to hundreds of billions of parameters~\cite{touvron2023llama}, incurs substantial computational costs of both processing power and memory requirements. 

Structured pruning is an effective way to deliver compact DNNs via identifying and removing redundant structures then recovering the lost knowledge~\cite{han2015deep,chen2021oto}. However, its application onto LLMs is facing significant challenges, due to the requirements of massive computational resources and the unavailable training datasets of both pretraining and instructed fine-tuning datasets~\citep{brown2020language}. Consequently, the paradigms could be largely categorized as pruning under \textit{limited} or \textit{full} resources.  For the limited-resource setup, recent pruning works~\cite{ma2023llm,zhang2023pruning,sun2023simple} uses Low-Rank-Adaptor (\lora)~\cite{hu2021lora} during either pruning and instructed fine-tuning stage to reduce the resource requirements, 
%with public datasets to seek knowledge recovery, 
yet still face significant performance degradation to the full \llm{}s. For the full-resouce setup, Sheared-LLaMA~\cite{xia2023sheared} conducts structured pruning on the original LLMs to directly achieve compact counterparts outperforming the equal sizes of LLMs trained from scratch, while requires significant GPU powers that might be not feasible for the public users. 

\begin{figure*}[t]
    \centering
    \includegraphics[width=0.95\linewidth]{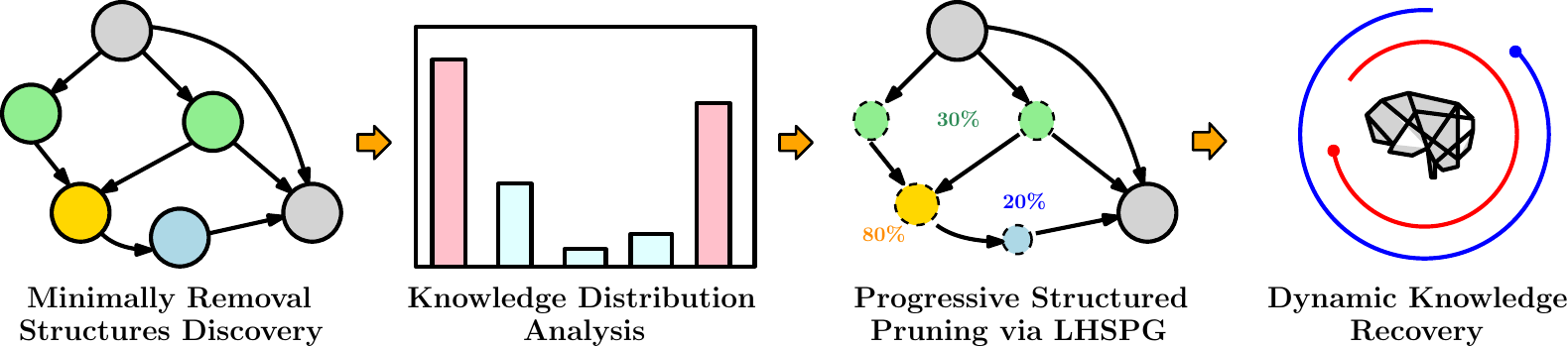}
    \caption{Overview of \algacro{}. Given a general LLM, \algacro{} at first discovers the minimally removal structures, then analyzes the knowledge distribution to mark the crucial ones as unprunable, then performs progressive structurally pruning over the prunable structures via \lorahspg{}, and finally recovers the lost knowledge to recap the performance gap to the full LLM.}
    \label{fig:overview}
\end{figure*}

We propose~\algacro{}, a novel structured pruning framework for LLMs in the limited-resource setup to significantly bring down the performance gap between pruned LLMs to their full versions. Compared with the existing works, \algacro{} has two main advantages to better preserve and recover the lost knowledge during pruning. Firstly, we proposed a novel Lora Half-Space Projected Gradient (\lorahspg) to enable progressive structured pruning with inherent knowledge transfer over \lora{} modules. Secondly, we propose a dynamic knowledge recovery stage to perform multi-stage fine-tuning in manner of both pretraining and instructed fine-tuning. Additionally, \algacro{} is applicable onto general LLMs with conducting dependency graph analysis over LLMs with LoRA modules upon~\cite{chen2023otov2}. 

We now summarize our main contributions as follows. 

\begin{itemize}[leftmargin=*]
    \item \textbf{Dependency Graph Analysis on LLMs with LoRA Modules.} To automatically structurally prune general LLMs, discovering the minimally removal structures is necessary via dependency graph analysis. LLMs with LoRA poses additional challenges, since target structures are non-trainable yet auxiliary LoRA modules are learnable, which are ignored by existing algorithms~\cite{chen2023otov2,ma2023llm}. We propose a novel graph algorithm to construct dependency graphs composed by overlapping node groups and composed node groups, and partition the trainable variables accordingly. 

    \item \textbf{Progressive Structured Pruning via \lorahspg{}.} We propose a novel structured sparsity optimization algorithm LoRA Half-Space Projected Gradient (\lorahspg{}) to perform progressive structured pruning. \lorahspg{} leverages the information from \lora{} modules and effectively produces desired structured sparsity over the original variables. \lorahspg{} transfers the knowledge stored in the relatively redundant structures to the important structures to better preserve the knowledge of the pretrained LLMs.
    
    \item \textbf{Dynamic Knowledge Recovery.} To further recover the knowledge after progressive pruning, we propose a dynamic knowledge recovery mechanism. Rather than only engaging the instructed fine-tuning as the existing limited-resource pruning works, we adaptively construct a subset from pretraining datasets upon the performance distribution to recover the lost general knowledge during pruning. We then perform the usual instructed fine-tuning to recover domain-specific expertise and the instruction capacity of pruned LLMs.
    
    \item \textbf{Experimental Results.} We demonstrate the effectiveness of \algacro{} on open-source \llamavone{}. By using one A100 GPU within a couple of GPU days, compared to the full model, the 20\% pruned \llamavone{} negligibly regresses 1\% performance, and the 50\% pruned \llamavone{} preserves 82\% performance on the evaluation benchmarks. Meanwhile, our results significantly outperform the existing state-of-the-arts.  

\end{itemize}

\section{Related Work}

While pruning~\cite{han2015deep} is well-established in traditional Deep Neural Networks (DNNs), its application to LLMs presents unique challenges. Unlike the smaller, task-specific DNNs~\cite{ding2021cdfi,ding2022sparsity}, LLMs have a large number of parameters and require significant computational resources~\cite{brown2020language}. Moreover, it's crucial for them to generalize well across multiple tasks~\cite{xia2023sheared}. Recently, various pruning methods have been developed specifically for LLMs, generally falling into two main categories: unstructured and structured.

\textbf{Unstructured Pruning.} Unstructured pruning methods~\cite{dong2017learning,chen2020neural,chen2021orthant} focus on setting unimportant individual weights in the model to zero. This fine-grained approach is straightforward and often maintains good performance, even with high compression rates. However, it results in sparse weight matrices that aren't well-suited for hardware accelerators, making them less efficient in real-world deployment. In the realm of LLMs, several new techniques have emerged. SparseGPT~\cite{sparsegpt} uses a sophisticated weight update process involving synchronized second-order Hessian updates, bypassing traditional retraining. In contrast, Wanda~\cite{sun2023simple} achieves high sparsity without any retraining, simply by pruning weights with the smallest magnitudes multiplied by their corresponding input activations. PST~\cite{pst}, however, combines unstructured pruning with efficient fine-tuning, pruning both LoRA and pre-trained model weights. A drawback of this method is the need for a memory-intensive mask that matches the shape of the pre-trained weights.

\textbf{Structured Pruning.} Structured pruning methods~\cite{chen2021oto,chen2023towards,chen2023otov2} focus on removing entire groups of parameters, such as neurons or layers, rather than individual weights. This group-level approach is hardware-friendly as it maintains dense weight matrices. The main challenge is selecting which structures to remove without compromising model performance. In the context of LLMs, several recent techniques aim for more efficient deployment and inference acceleration. For example, LLM-Pruner~\cite{ma2023llm} proposes a dependency detection algorithm to identify and remove non-critical coupled structures, followed by a rapid post-training phase with limited data. However, this method is memory-intensive as it requires full gradient information and is not compatible with LoRA, necessitating a separate post-training phase for knowledge recovery. In contrast, LoRAPrune~\cite{zhang2023pruning} integrates LoRA with iterative structured pruning, achieving both parameter-efficient fine-tuning and direct hardware acceleration. This approach is also memory-efficient, relying only on LoRA's weights and gradients for pruning criteria, unlike LLM-Pruner, which uses full gradients. Most recently, Sheared-LLaMA~\cite{xia2023sheared} aims to prune the model to a target architecture defined by existing pre-trained models. It then trains the pruned model using dynamically loaded data, based on each domain's rate of loss reduction, leading to more efficient data usage and faster performance improvement. However, Sheared-LLaMA allocates considerable computational resources to subsequent pre-training for performance recovery.

In this work, we present LoRAShear, a method for efficient structured pruning of LLMs while recovers knowledge. Compared to the existing methods, our approach uniquely leverages a novel graph algorithm to create dependency graphs for both the original LLM \emph{and} LoRA modules. We further introduce a structured sparsity optimization algorithm that utilizes information from LoRA modules to update weights, thereby enhancing knowledge preservation. Following pruning, we employ a dual-phase training approach involving both pre-training and fine-tuning to recover general and domain-specific knowledge effectively.

\section{\algacro{}}

\algacro{} dedicately designs a comprehensive end-to-end pipeline to compress pretrained LLMs and deliver efficient knowledge recovery. The outlined is stated as Algorithm~\ref{alg.main.outline}. Given a general LLM $\mathcal{M}$, we at first analyze its architecture,  create its dependency graph, and partition its trainable variables into a group set $\mathcal{G}$ following the discovered minimally removal structures (Section~\ref{sec.minimal_removal_structures}). We then analyze the knowledge distributed over the minimally removal structures to exclude the ones that highly impact the model performance from pruning (Section~\ref{sec.knowledge_distribution_analysis}). Next, progressive structured pruning is performed over the prunable structures via our proposed \lorahspg{} to identify redundant  structures and transfer the knowledge stored in the redundant structures back onto the important counterparts (Section~\ref{sec.lhspg}), and construct a compressed LLM $\mathcal{M}^*$ via automatically removing redundant structures (Section~\ref{sec.compression}). The lost knowledge during pruning is then recovered via a dynamic knowledge recovery stage to recap the performance of the compressed  $\mathcal{M}^*$ to the full LLM (Section~\ref{sec.dynamic_fine_tuning}).

\begin{algorithm}[h!]
\caption{Outline of \algacro{}.}
\label{alg.main.outline}
\begin{algorithmic}[1]
    \STATE \textbf{Input.} A general pretraining LLM $\mathcal{M}$. 
    \STATE \textbf{Discover minimal removal structures} of $\mathcal{M}$ via creating and analyzing dependency graph $(\mathcal{V}, \mathcal{E})$. Partition trainable variables of $\mathcal{M}$ into $\mathcal{G}$.
    \STATE \textbf{Analyze knowledge distribution} over each node group in the dependency graph. 
    \STATE \textbf{Progressive structured pruning by \lorahspg} to identify redundant structures and transfer lost knowledge.
    \STATE \textbf{Construct compressed model} to erasing redundancy to form compressed compact LLM $\mathcal{M}^*$.
    \STATE \textbf{Dynamic fine-tuning} to recover lost knowledge.
    \STATE \textbf{Output.} The compact high-performing LLM $\mathcal{M}^*$.
\end{algorithmic}
\end{algorithm}

\begin{figure*}[t]
\begin{subfigure}[b!]{\textwidth}
\centering
\includegraphics[width=0.85\linewidth]{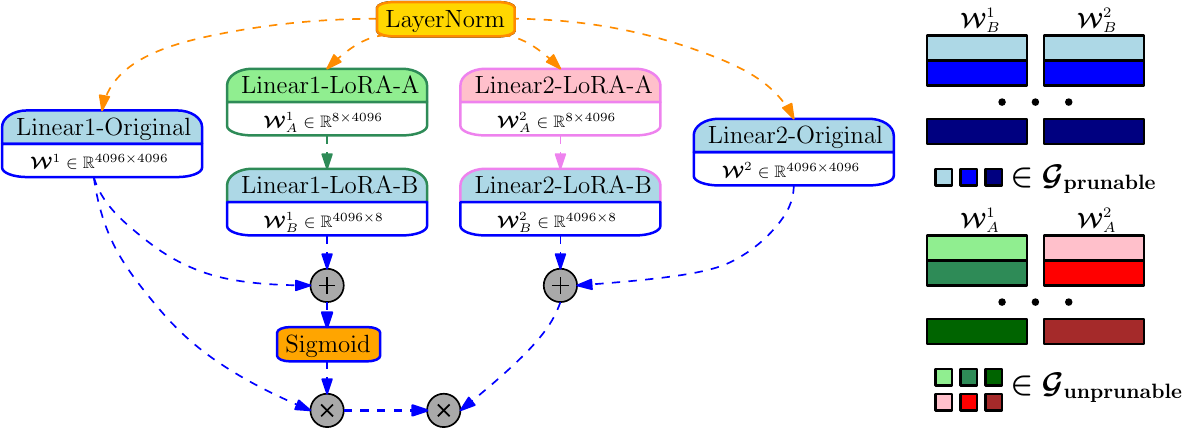}
\caption{Dependency graph for MLP layers.}
\label{fig:dependancy_mlp}
\end{subfigure}
\begin{subfigure}[b!]{\textwidth}
\centering
\includegraphics[width=0.85\linewidth]{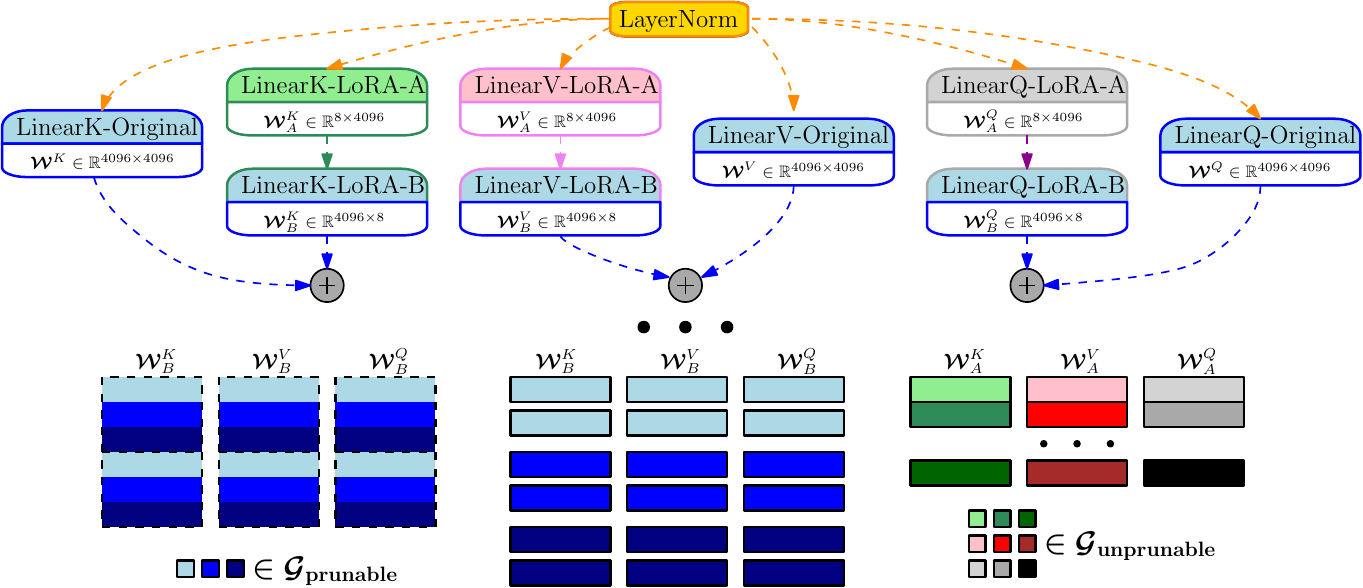}
\caption{Dependency graph for Attention layers.}
\label{fig:dependancy_attn}
\end{subfigure}    
\caption{Dependency graph in \llamavone{} and trainable variable partitions.}
\label{fig:dependancy}
\end{figure*}

\subsection{Minimally Removal Structure Discovery}\label{sec.minimal_removal_structures}

\begin{algorithm}[h!]
\caption{Minimally Removal Structure Discovery.}
\label{alg.main.dmrs}
\begin{algorithmic}[1]
\STATE \textbf{Input.} A LLM $\mathcal{M}$ to be compressed and fine-tuned.
\STATE Construct the trace graph $(\mathcal{E}, \mathcal{V})$ of $\mathcal{M}$.
\STATE Establish node groups $\mathcal{N}_\text{composed}$ for composed operators via traversing $(\mathcal{E}, \mathcal{V})$ and the module tree of $\mathcal{M}$.\label{line.composed_node_groups}
\STATE Establish node groups $\mathcal{N}_\text{basic}$ for remaining operators.
\STATE Build dependancy across $\mathcal{N}_\text{composed}$ and $\mathcal{N}_\text{basic}$.
\STATE Partition trainable variables into minimally removal structures and form $\mathcal{G}$.
\STATE \textbf{Return} the trainable variable groups $\mathcal{G}$.
\end{algorithmic}
\end{algorithm}

Given a target LLM $\mathcal{M}$, the foremost step is to discover the minimally removal structures, which are defined as the units that can be directly removed without affecting the functionality of the remaining DNNs. Such discovery was achieved by analyzing the trace graphs and creating  the dependency graphs over the basic operators in OTOv2~\cite{chen2023otov2}. In these dependency graphs, each node group indicates the operators that are dependent and needs to be pruned together if with trainable variables and are disjoint to each other in the normal DNNs. However, LLMs with LoRA modules easily disrupt those algorithms since in such models, only LoRA modules are trainable, and the original LLM variables are fixed yet prunable.  To address issue, we dedicately introduce composed operator and overlapping node groups. Composed operator refers to the operators that are assembled by multiple basic operators such as LoRA modules consisting of two \texttt{linear} operators, yet needs to be considered as an entirety. Each such composed operator should form one node group, such as \texttt{Linear-1-LoRA-A-Linear-1-LoRA-B} in Figure~\ref{fig:dependancy_mlp}. The overlapping node groups $\mathcal{N}_\text{composed}$ exist because their outgoing nodes still need to obey the dependency across other adjacent operators in the trace graph, \eg, \texttt{Linear-1-LoRA-B} belonging to two node groups marked as \textcolor{green}{green} and \textcolor{blue}{blue} simultaneously. We then jointly consider basic and composed node groups $\mathcal{N}_\text{basic}$ and $\mathcal{N}_\text{composed}$ and partition the trainable variables of $\mathcal{M}$ into a set of groups $\mathcal{G}$, wherein each group $g\in\mathcal{G}$ corresponds to one minimally removal structure.

\subsection{Knowledge Distribution Analysis}\label{sec.knowledge_distribution_analysis}

\begin{algorithm}[h!]
    \caption{Knowledge Distribution Analysis.}
    \label{alg.main.knowledge.distribution.analysis}
    \begin{algorithmic}[1]
        \STATE \textbf{Input.} Trainable variable partition $\mathcal{G}$, node groups $\mathcal{N}_\text{composed} \cup \mathcal{N}_\text{basic}$, a set of pruning ratios $\mathcal{P}$, an evaluation dataset $\mathcal{D}_\text{eval}$, and a unprunable ratio $\gamma$. 
        \FOR{each node group in $\mathcal{N}_\text{composed} \cup \mathcal{N}_\text{basic}$}
            \STATE Prune groups upon some specified proxy and $\mathcal{P}$.
            \STATE Compute performance deviation upon $\mathcal{D}_\text{eval}$.
            \STATE Recover pruned groups to the original status.
        \ENDFOR
        \STATE Sort the perform deviation over $\mathcal{N}_\text{composed} \cup \mathcal{N}_\text{basic}$.
        \STATE Mark the groups in $\mathcal{G}$ regarding the node groups with the largest deviation upon $\gamma$ as unprunable $\mathcal{G}_\text{unprunable}$.
        \STATE Mark the remaining groups in $\mathcal{G}$ as prunable $\mathcal{G}_\text{prunable}$.
        \STATE \textbf{Return}  prunable and unprunable variable groups $\mathcal{G}_\text{prunable}\cup \mathcal{G}_\text{unprunable}$.
    \end{algorithmic}
\end{algorithm}

Due to the universal training process, the knowledge is unevenly distributed across all the node groups in the dependency graph. Some node groups serve remarkably more significant roles than  others, resulting in performance collapse if pruning them. Meanwhile, in the limited resources setting, the knowledge after collapse would not be easily recovered. Therefore, before engaging into the progressive structured pruning stage, we analyze the knowledge distribution to locate the node groups that should be excluded from pruning. As stated in Algorithm~\ref{alg.main.knowledge.distribution.analysis}, we iteratively traverse all node groups, and prune each of them upon some specified pruning ratio yet keep the remaining groups unchanged. We then evaluate the output deviation between each pruned LLM against the full model upon some pre-selected evaluation dataset. The ones with the largest $\gamma |\mathcal{N}_\text{composed}\cup \mathcal{N}_\text{basic}|$ deviations are marked as unprunable, which corresponding groups of variables to form $\mathcal{G}_\text{unprunable}$. The remaining ones are marked as prunable, where trainable variables form $\mathcal{G}_\text{prunable}$.

\subsection{Progressive Structured Pruning via LHSPG}\label{sec.lhspg}

The next step is to proceed progressive structured pruning over the prunable groups of variables $\mathcal{G}_\text{prunable}$. To proceed it, we propose a novel structured sparsity optimizer LoRA Half-Space Projected Gradient (\lorahspg) to yield structured sparsity over the original model parameters based on the optimization information over auxiliary LoRA modules. There exist two main takeaways of \lorahspg{}, \ie, \textit{(i)} effectively identify and remove redundant structures via projecting them onto zero, and \textit{(ii)} transfer the knowledge stored in the relatively redundant structures to be pruned back to the important counterparts to better preserve the knowledge of full LLMs.  

\begin{figure}
    \centering
    \includegraphics[width=\linewidth]{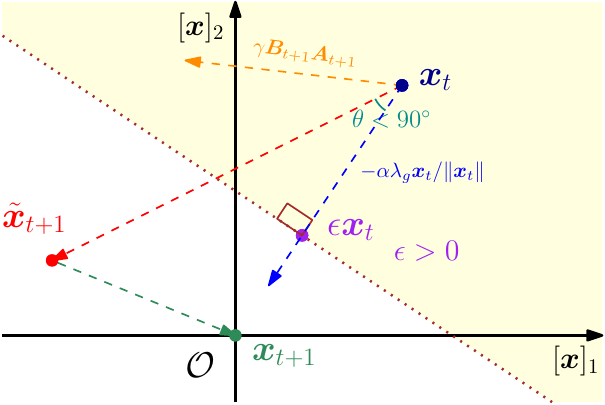}
    \caption{Half-Space step over LoRA modules. }
    \label{fig:enter-label}
\end{figure}

\begin{algorithm}[h!]
\caption{Progressive Structured Pruning via LHSPG}
\label{alg:main.dhspg}
\begin{algorithmic}[1]
\STATE \textbf{Input.} pretraining variable $\bm{x}_0$, learning rate $\alpha$, warm-up steps $T_w$, progressive period $P$, period length $T_p$, target group sparsity level $K$, and variable partition $\mathcal{G}_\text{prunable}\cup \mathcal{G}_\text{unprunable}$.
\STATE Warm up $T_w$ steps via SGD or its variants (AdamW).\label{line:warm_up}
\STATE Initialize redundant groups $\mathcal{G}_\text{redundant}\gets \emptyset$.
\STATE Initialize important groups $\mathcal{G}_\text{important}\gets \mathcal{G}$.
\STATE Compute sparsity level for each pruning period $\widehat{K}:=K/T_p$.\label{line:sparsity_level_period}
\FOR{each pruning period $p=0, 1, \cdots, P-1$ }
    \STATE Pickup $\widehat{\mathcal{G}}_{p}$ in $\mathcal{G}_\text{important}$ with $\widehat{K}$-least saliency scores.\label{line:pickup_least_important}
    \STATE Update $\mathcal{G}_\text{redundant}\gets \mathcal{G}_\text{redundant}\cup \widehat{\mathcal{G}}_{p}$.\label{line:update_redundant_groups}
    \STATE Update $\mathcal{G}_\text{important}\gets \mathcal{G}_\text{important}/\widehat{\mathcal{G}}_{p}$.\label{line:update_important_groups}
    \FOR{$t=0, 1, \cdots, T_p-1$}
        \STATE Update LoRA $\bm{B}$ and $\bm{A}$ via SGD or its variants.
        \begin{equation}
        \begin{split}
        \bm{B}_{t+1}\gets \bm{B}_{t} - \alpha_k \Grad_{\bm{B}_{t}} f\\     
        \bm{A}_{t+1}\gets \bm{A}_{t} - \alpha_k \Grad_{\bm{A}_{t}} f        
        \end{split}
        \end{equation}
        \STATE Compute trial iterate $[\tilde{\bm{x}}_{t+1}]_{\widehat{\mathcal{G}}_{p}}$ for each $g\in \widehat{\mathcal{G}}_{p}$.
        \begin{equation}
        \ \ \ \ \ \ \ \ \ \ [\tilde{\bm{x}}_{t+1}]_{g}\gets[\bm{x}_t+\gamma \bm{B}_{t+1}\bm{A}_{t+1}]_g-\frac{\lambda_g[\bm{x}_t]_g}{\norm{[\bm{x}_t]_g}}
        \end{equation}
        \STATE Perform Half-Space projection over $[\tilde{\bm{x}}_{t+1}]_{\widehat{\mathcal{G}}_{p}}$.
        \STATE Update $[\bm{x}_{t+1}]_{\widehat{\mathcal{G}}_{p}}\gets [\tilde{\bm{x}}_{t+1}]_{\widehat{\mathcal{G}}_{p}}$.
        \STATE Update $[\bm{B}_{t+1}]_{\widehat{\mathcal{G}}_p}\gets \bm{0}$.
        \IF{$t=T_p-1$}
            \STATE Merge $[\bm{B}_{t+1}\bm{A}_{t+1}]_{\mathcal{G}_\text{important}}$ into $[\bm{x}]_{\mathcal{G}_\text{important}}$.
        \ENDIF 
    \ENDFOR
\ENDFOR
\STATE \textbf{Return} the final iterate $\bm{x}^*_{\lorahspg}$.
\end{algorithmic}
\end{algorithm}

\textbf{Target Problem.} We formulate the progressive structured pruning as the following structured sparsity optimization problem~\eqref{prob.main} over LLMs with LoRA modules. 
\begin{equation}\label{prob.main}
\minimize{\mathcal{A}, \mathcal{B}}\ f(\mathcal{A}, \mathcal{B}),\ \ \text{s.t.} \ \text{Card}\{g\in\mathcal{G}_\text{prunable}| [\bm{x}]_g=0\}= K,
% \vspace{-2mm}
\end{equation}
where  $\mathcal{A}$ and $\mathcal{B}$ are the collections of LoRA decomposing sub-matrices, which are trainable during the structured pruning. We seek to yield group sparsity over the original variables with the target sparsity level as $K$.

\textbf{Outline.} The outline of \lorahspg{} is presented in Algorithm~\ref{alg:main.dhspg}. We at first warm up the LoRA variables in the prunable groups $\mathcal{G}_\text{prunable}$ via stochastic gradient descent (SGD) or its variants like AdamW to collect gradient information. We then progressively identify redundant structures within $P$ periods of sparse optimization. To proceed, we compute the target group sparsity level to be produced for each period. In each period $p$, we sort the prunable groups upon some prespecified saliency proxies and pick up the ones with least saliency scores as redundant groups for the current period $\widehat{\mathcal{G}}_p$. Then we compute trial iterate over the LoRA variables in $\mathcal{A}$ and $\mathcal{B}$ via SGD or its variants. For the redundant groups $\widehat{\mathcal{G}}_p$, we proceed a gradient descent via LoRA approximation and penalize over the variable magnitude proportionally to $\lambda_g$, which is selected upon the length of each pruning period. A Half-Space projection is next performed over the trial iterate to project groups of variables with the least sacrificing over the objective function. During the whole process, redundant groups are progressively projecting onto zero, during the projection, the LoRA modules for the important counterparts are absorbing the knowledge via minimizing the loss functions. As a result, the progressive structured pruning not only effectively identifies and projects redundant groups of variables onto zero, but also preserve the knowledge stored in the redundant structures to the largest extent. A final iterate $\bm{x}_\text{LHSPG}^*$ is returned for the subsequent step. 

\subsection{Compressed LLM Construction}\label{sec.compression}

Given the solution of \lorahspg{}, \algacro{} automatically constructs a structurally pruned LLM $\mathcal{M}^*$ via automatically erasing the structures corresponding to the redundant groups in $\mathcal{G}_\text{prunable}$. The whole procedure is performed via two pass dependency graph traversal. The first-pass traversal iterates each node group and prunes the structures along the primary dimension. The second-pass traversal erases the structures along the secondary dimension upon the pruned status of the incoming structures.

\subsection{Dynamic Knowledge Recovery}\label{sec.dynamic_fine_tuning}

\begin{algorithm}[h!]
    \caption{Dynamic Knowledge Recovery.}
    \label{alg.main.knowledge.recovery}
    \begin{algorithmic}[1]
        \STATE \textbf{Input.} pretraining dataset~$\mathcal{D}_\text{pretraining}$, instructed fine-tuning dataset $\mathcal{D}_\text{instruct}$, and a pruned LLM $\mathcal{M}^*$.
        \STATE Establish validation datasets for $\mathcal{D}_\text{pretraining}$ and $\mathcal{D}_\text{instruct}$ as $\mathcal{D}_\text{pretraining}^{\text{val}}$ and $\mathcal{D}_\text{instruct}^{\text{val}}$, respectively.
        \FOR{$\mathcal{D}\in \{\mathcal{D}_\text{pretraining}, \mathcal{D}_\text{instruct}\}$}
            \WHILE{not converge}
                \STATE Dynamically construct $\widehat{\mathcal{D}}\subseteq \mathcal{D}$ by evaluation.
                \STATE Fine-tune $\mathcal{M}^*$ with LoRA on $\widehat{\mathcal{D}}$.
            \ENDWHILE
        \ENDFOR
        \STATE \textbf{Return} knowledge-recovered pruned LLM $\mathcal{M}^*$.
    \end{algorithmic}
\end{algorithm}

The final step is recovering lost knowledge after pruning and restoring the capabilities of LLM. To achieve successful recovery, it's essential to understand how the LLM acquires its knowledge. The knowledge is acquired through a two-stage process: pretraining on extensive and diverse text corpora, followed by fine-tuning with specific instruction datasets. The acquired knowledge is stored as variables within the LLM, but these variables are removed during the pruning process. Therefore, to regain the knowledge, a post-training process is required, involving both the pretraining and instructed fine-tuning datasets.

Due to the vast and diverse nature of the pretraining datasets, existing structured pruning methods, especially those designed for limited resources, only rely on the instructed fine-tuning datasets. However, this approach often leads to a significant degradation in performance. To mitigate this challenge, we introduce a dynamic knowledge recovery framework, presented as Algorithm~\ref{alg.main.knowledge.recovery}.

In particular, given the pretraining and the instructured fine-tuning dataset collections $\mathcal{D}_\text{pretraining}$ and $\mathcal{D}_\text{instruct}$, we at first uniformly sample subsets from them for validation $\mathcal{D}_\text{pretraining}^{\text{val}}$ and $\mathcal{D}_\text{instruct}^{\text{val}}$. We then consider knowledge recovery over pretraining datasets. To proceed, we at first evaluate the performance degradation over the different sources via $\mathcal{D}_\text{pretraining}^{\text{val}}$. Upon on the performance deviation distribution, we construct a subset $\widehat{\mathcal{D}}_\text{pretraining}\subseteq \mathcal{D}_\text{pretraining}$.
The criteria for selecting samples involve prioritizing categories experiencing more significant degradation while ensuring a balanced representation of samples from sources with minimal degradation to prevent overfitting. Subsequently, we employ LoRA for fine-tuning the pruned model. If the evaluation results do not converge, we repeat the process of constructing the next subset from $\mathcal{D}_\text{pretraining}$ until convergence is achieved. Following the knowledge recovery from the pretraining stage, we apply the same methodology to the instructed fine-tuning datasets. This iterative approach ultimately yields the highly optimized pruned LLM $\mathcal{M}^*$. 

\section{Numerical Experiments}

To demonstrate the effectiveness of \algacro{}, we provide our preliminary results on open-source \llamavone{} \cite{touvron2023llama}. More experimental results will come in the future revisions. 

\subsection{Dataset Selection}

\textbf{Pretraining Datasets.} We follow~\citeauthor{touvron2023llama} to collect pretraining datasets or the alternatives for English. In particular, we select the OpenWebText~\cite{Gokaslan2019OpenWeb} as an alternative to English CommonCrawl and C4 datasets. We select a processed Wikipedia dump on 2022 \cite{wikidump}. Gutenberg~\cite{gerlach2020standardized} and BookCorpus~\cite{Zhu_2015_ICCV} are also used in our collection. For each datasets, we proceed standard pre-processing to erase irregular characters and only keep the paragraphs that contains more than 64 tokens. 

\textbf{Instructed Fine-Tuning Datasets.} For fair comparison, we follow the existing structured pruning LLM works~\cite{ma2023llm,zhang2023pruning} in the limited-resource setting to use the Alpaca dataset~\cite{alpaca}, which consists 52,000 instructions and demonstrations generated by OpenAI's text-davinci-003 engine. Alpaca is frequently used to conduct general instruction-tuning for language models and make the LLM follow instruction better.

\subsection{Experimental Results}

\begin{figure*}[t]
    \centering
    \includegraphics[width=\linewidth]{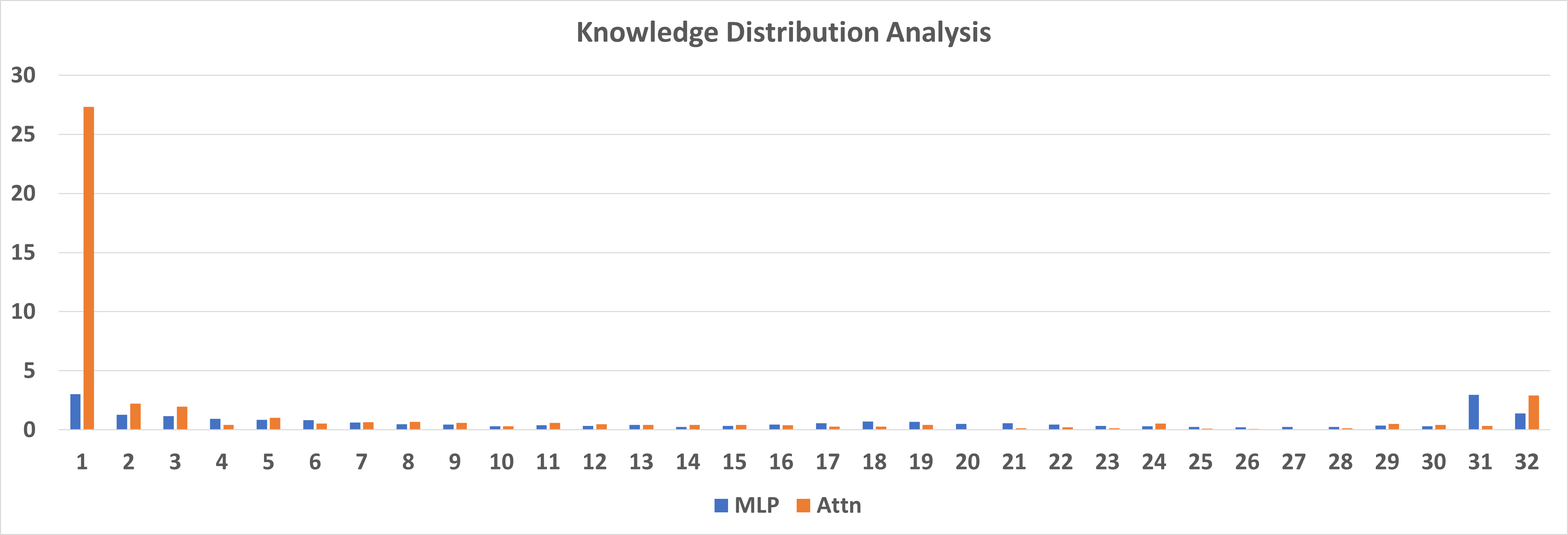}
    \vspace{-7mm}
    \caption{Knowledge distribution analysis by measuring the perplexity deviation to the full \llamavone{}.}
    \label{fig:knowledge_distribution}
\end{figure*}
% \vspace{-4mm}
\begin{table*}[t]
\centering
\vspace{-4mm}
\caption{\algacro{} over \llamavone{}.}\label{table.comp_llamav1}
\vspace{2mm}
     % \scriptsize
\resizebox{\linewidth}{!}{
\begin{tabular}{ll|ccccccc|c}
	\toprule 
    Pruning Ratio & Method & BoolQ & PIQA & HellaSwag & WinoGrande & ARC-e & ARC-c & OBQA & Average \\
	\hline
    Ratio $= 0\%$ & \llamavone{}~\cite{touvron2023llama} & 76.5 & 79.8 & 76.1 & 70.1 & 72.8 & 47.6 & 57.2 & 68.59 \\
    (Baseline) & \llamavone{}~\cite{ma2023llm} & 73.18  & 78.35 & 72.99 & 67.01 & 67.45 & 41.38 & 42.40 & 63.25 \\
    % \hdashline
    % Baseline & \llamavone{}~\cite{ma2023llm} & 73.18  & 78.35 & 72.99 & 67.01 & 67.45 & 41.38 & 42.40 & 63.25 \\
    \hdashline
    Ratio $= 20\%$ & LLM-Pruner~\cite{ma2023llm} & 66.79 & 77.58 & 68.48 & 64.96 & 64.06 & 37.88 & 39.00 & 59.82 \\
    & LoRAPrune~\cite{zhang2023pruning} & 65.82 & 79.31 & 70.00 & 62.76 & 65.87 & 37.69 & 39.14 & 60.05 \\ 
    & WANDA~\cite{sun2023simple} & 65.75 & 74.70 & 64.52 & 59.35 & 60.65 & 36.26 & 39.40 & 57.23 \\
    & \textbf{\algacro}$^\dagger$ & 70.17 & 76.89 & 68.69 & 65.83 & 64.11 & 38.77 & 39.97 & 60.63 \\
    & \textbf{\algacro} & 72.78 & 76.36 & 69.49 & 67.63 & 69.02 & 39.47 & 40.78 & \textbf{62.22} \\
    \hdashline
    Ratio $= 50\%$ & LLM-Pruner~\cite{ma2023llm} & 61.56 & 68.72 & 46.62 & 52.64 & 47.94 & 29.27 & 35.40 & 48.88 \\
    & LoRAPrune~\cite{zhang2023pruning} & 61.88 & 71.53 & 47.86 & 55.01 & 45.13 & 31.62 & 34.98 & 49.71  \\ 
    & WANDA~\cite{sun2023simple} & 50.90 & 57.38 & 38.12 & 55.98 & 42.68 & 34.20 & 38.78 & 45.43\\ 
    & \textbf{\algacro{}}$^\dagger$ & 62.12 & 71.80 & 48.01 & 56.29 & 47.68 & 32.26 & 34.61 & 50.39 \\
    & \textbf{\algacro{}} & 63.40 & 72.15 & 49.83 & 56.40 & 49.45 & 34.31 & 35.86 & 51.63 \\
    \bottomrule
    \multicolumn{10}{l}{$^\dagger$ Knowledge recovery only on the instructured fine-tuning datasets as other works.}\\ % 0514-2atm
    % \multicolumn{10}{l}{$^\dagger$ .}
\end{tabular}
}
\end{table*}

\textbf{Knowledge Distribution Analysis.}  The analyzed knowledge distribution on \llamavone{} is presented in Figure~\ref{fig:knowledge_distribution}. Given an evaluation dataset, we perform Algorithm~\ref{alg.main.knowledge.distribution.analysis} to analyze the knowledge distributed across the minimally removal structures in each node group. After measuring the output deviation, it is apparent that the knowledge is unevenly distributed across different node groups. The first and last few node groups serve as more significant roles than others to the model prediction. During pruning, it would be better to avoid pruning these most sensitive node groups since the saliency score calculation may still prune some of their minimally removal structures which may result in significant performance degradation.

\textbf{Pruning Results.} We now show the quantitative results of \algacro{} and compare with other methods over the evaluation benchmark computed via \texttt{lm-evaluation-harness} \cite{eval-harness}.  As shown in Table~\ref{table.comp_llamav1}, under the same pruning ratio 20\%, \algacro{} significantly outperforms others by 2.2\%-5.0\% accuracy and negligibly regress 1\% compared to the full \llamavone{}. We additionally conduct an ablation that only levering the same instructured fine-tuning dataset, \ie, Alpaca, to recover the lost knowledge. The performance in this setting still outperform other methods, implying the effectiveness of progressive structured pruning via \lorahspg{} to transferring and preserving knowledge. Under the high pruning ratio 50\%, the outerperformance of \algacro{} still holds. In particular, under both progressive structured pruning followed by knowledge recovery via pretraining and instructured fine-tuning datasets, our performance is significantly better than the existing state-of-the-arts.  

\section{Conclusion}

We propose a novel \algacro{} to conduct efficient structured pruning and knowledge recovery for general LLMs in the limited resources setup. \algacro{} has three takeaways: \textit{(i)} it automatically discovers minimally removal structures over LLMs with LoRA modules; \textit{(ii)} conducts progressive structured pruning via a novel structured sparsity optimizer \lorahspg{} that yields structured sparsity over original variables via the information stored in LoRA modules; and \textit{(iii)} equips with a dynamic knowledge recovery stage to gain knowledge back from both pretraining and instructured fine-tuning datasets. Numerical results validates the efficacy, that negligibly regress 1\% performance to the full model under 20\% pruning ratio and preserve 82\% performance under 50\% pruning ratio to the full LLMs. More experiments will come in the updated versions. 

\bibliography{icml2024_paper}
\bibliographystyle{icml2023}

\end{document}